\documentclass[a4paper, twocolumn, 10pt]{article}

\usepackage[left=1.5cm, right=1.5cm, top=2cm, bottom=2cm]{geometry}

\usepackage{authblk}

\usepackage{graphicx}
\usepackage{amsmath}
\usepackage{amssymb}
\usepackage{hyperref}
\usepackage{booktabs} 
\usepackage{multirow} 
\usepackage{pifont}
\usepackage{caption}

\usepackage{titlesec}
\titleformat*{\section}{\Large\bfseries\raggedright}
\titleformat*{\subsection}{\large\bfseries\raggedright}
\titleformat*{\subsubsection}{\normalsize\bfseries\raggedright}

\title{Graph-Augmented Topological Internalization with Dual-Stream Classifiers for Medical Report Generation}

\author[1]{Moyu Tang}
\author[1,\dag]{Shangkun Sima}
\author[1,\dag]{Chupei Tang}
\author[1,\dag]{Junxiao Kong}
\author[1,\dag]{Di Wang}
\author[2,*]{Tianchi Lu}

\affil[1]{\small School of Mathematics and Statistics, Lanzhou University, 222 South Tianshui Road, 730000, Gansu, China.}
\affil[2]{\small Department of Computer Science, City University of Hong Kong, 83 Tat Chee Avenue, Kowloon Tong, Hong Kong 999077, China.}

\date{} 

\begin{document}

\maketitle

\begingroup
\renewcommand{\thefootnote}{\fnsymbol{footnote}}
\footnotetext[2]{Equal contribution. These authors should be considered co-second authors.}
\footnotetext[1]{Corresponding author: tianchilu4-c@my.cityu.edu.hk; Tel: +86-13239620274.}
\endgroup

\begin{abstract}
Automated medical report generation, MRG, holds substantial value for alleviating radiologist workload and enhancing diagnostic efficiency. However, mainstream approaches typically treat diverse chest abnormalities as isolated classification targets. This paradigm often overlooks inherent disease co-occurrences and struggles to translate medical topological structures into explicit data correlations, constraining the model's reasoning capacity on complex or subtle lesions. To address this, we propose a Graph-Augmented Dual-Stream Medical Report Generation with Topological Internalization, GDMRG. Our framework introduces a Topological Knowledge Internalization module, TKI, which leverages a Graph Convolutional Network, GCN, to generate an explicit parameterized weight matrix based on global disease co-occurrence priors. This facilitates efficient topological knowledge injection without relying on external retrieval mechanisms. Building upon this, we construct a dual-stream classification system: the main branch generates discrete diagnostic prompts under topological constraints, while the auxiliary branch employs an asymmetric optimization strategy to dynamically calibrate decision boundaries for highly imbalanced samples. Concurrently, to establish a logical closed loop between diagnosis and visual grounding, we design a diagnostic-driven Diagnosis-Guided Spatial Attention, DGSA, that utilizes high-dimensional clinical semantics to recalibrate the visual encoder, mitigating feature hallucinations. Comprehensive experiments on the MIMIC-CXR dataset demonstrate that GDMRG achieves competitive clinical efficacy, CE, while maintaining natural language fluency. Furthermore, our model exhibits robust zero-shot generalization on the IU X-Ray dataset. In summary, this work presents an integrated and interpretable paradigm for medical report generation.
\end{abstract}

\vspace{0.5cm}

\section{Introduction}

Automatic Medical Report Generation (AMRG) aims to translate complex medical images into professional, structured natural language descriptions~\cite{jing2018automatic, chen2020generating}. This task holds substantial clinical significance for alleviating radiologist workload and improving diagnostic efficiency~\cite{brady2017error}. However, clinical pathology indicates that chest abnormalities rarely occur in isolation~\cite{yu2020c2fnas}; rather, they exhibit strong anatomical and complication correlations (e.g., pleural effusion is often accompanied by cardiomegaly). While current deep learning models have made progress, they typically treat multi-label disease classification as independent tasks~\cite{liu2019clinically}. These approaches capture disease correlations primarily through implicit representations within the deep feature space. Lacking explicit medical topological constraints, such mechanisms often struggle to fully model the prior logic of clinical diagnosis and the intrinsic statistical relationships among diseases. Consequently, they face reasoning challenges when dealing with subtle features or complex complications.

To address this limitation, we propose a Graph-Augmented Dual-Stream Medical Report Generation with Topological Internalization (GDMRG). Our primary objective is to transform implicit medical co-occurrence associations into explicit, data-level correlations~\cite{chexgcn}. Specifically, GDMRG introduces a Topological Knowledge Internalization (TKI) module, which constructs a static adjacency matrix utilizing global co-occurrence statistical patterns from medical datasets alongside word embeddings~\cite{pennington2014glove}. Through the internal Graph Convolutional Network (GCN) architecture of the TKI, the model incorporates this explicit disease topology into the parameter space of the classifier. This mechanism enables diagnostic decisions to be supported by explicit topological representations, thereby achieving efficient topological knowledge injection without relying on resource-intensive external retrieval mechanisms.

To bridge the heterogeneous semantic gap between visual representations and structured text, recent advancements in vision-language alignment have explored prompt-driven mechanisms~\cite{zhou2022learning, liu2023pre}. Within the AMRG domain, an emerging strategy formulates discrete diagnostic labels as prefix prompts, which explicitly guide downstream language models to auto-regressively generate reports~\cite{promptmrg}. While this discrete alignment approach improves cross-modal feature grounding, its effectiveness remains constrained by the accuracy of the upstream diagnostic predictions. Furthermore, existing methods for generating these discrete prompts predominantly rely on standard independent linear classifiers~\cite{promptmrg, miura2021improving}.

To better adapt the prompt-driven paradigm to complex clinical scenarios, we enhance the prompt generation mechanism via a graph-augmented dual-stream classification system. In the main classification branch, we apply a weight-tying mechanism using the topological weights generated by the TKI module, moving beyond standard independent predictions. Rather than overfitting isolated sparse samples in a long-tailed distribution, this mechanism acts as a structural regularizer that constrains the classifier's hypothesis space. By trading off a marginal recall rate associated with low-confidence predictions, it effectively reduces false positives (FPs) that lack pathological support. This topological constraint encourages the model to yield predictions that better align with clinical priors, particularly for rare complications. Concurrently, the auxiliary branch introduces a domain-adapted Truncated Asymmetric Loss (T-ASL)~\cite{ridnik2021asymmetric}. By dynamically decoupling the gradients of abundant simple negative samples, T-ASL mitigates the impact of medical label noise via a bidirectional gradient truncation mechanism. This parameterized dual-stream joint optimization maintains architectural cohesion while providing clinically reliable prior guidance for the subsequent text decoder.

In summary, the main contributions of this work are as follows:
\begin{itemize}
    \item We propose GDMRG, a graph-augmented dual-stream classification framework. By transforming implicit disease associations into parameterized explicit topological constraints via the TKI module, our framework achieves structural knowledge injection without relying on external cross-modal retrieval mechanisms.
    \item To address the challenges of long-tailed missed diagnoses and feature hallucinations in prompt-driven paradigms, we design a dual-classifier system integrating topological regularization, T-ASL, and an optimal threshold adaptive strategy. This design enhances diagnostic reliability under class imbalance and label noise.
    \item We introduce a Diagnosis-Guided Spatial Attention (DGSA) mechanism, establishing a logical feedback loop from medical diagnosis to lesion visual grounding. This module improves the model's interpretability and the factual fidelity of the generated text.
    \item Comprehensive experiments demonstrate that GDMRG achieves competitive performance against existing methods under two standard clinical efficacy (CE) evaluation protocols, offering a cohesive and interpretable paradigm for medical report generation.
\end{itemize}

\section{Related Work}

\subsection{Medical Report Generation}
Automated medical report generation (MRG) has evolved from early CNN-RNN architectures~\cite{wang2018tienet, xue2018multimodal, li2018hybrid} to complex Transformer-based feature alignment models~\cite{chen2020generating, miura2021improving,  rui2021artificial}. Recently, to bridge the heterogeneous gap between visual representations and structured medical text, prompt-driven generation strategies have been introduced to the field~\cite{jia2021visual, promptmrg}. These approaches typically extract discrete disease labels to serve as diagnostic prefixes (prompts), which subsequently constrain and guide the language model during text decoding to improve the clinical accuracy of the reports~\cite{liu2023pre, zhou2022learning}. The viability of this mechanism demonstrates the fundamental value of discrete medical concepts in bridging cross-modal generation. Building upon this foundational structural design, our study investigates a critical bottleneck: how to generate precise diagnostic prompts from real-world clinical data characterized by class imbalance and complex complications.

\subsection{Knowledge Graphs and Topology in Medical Imaging}
In clinical diagnosis, chest abnormalities are often accompanied by complex anatomical complications~\cite{yu2020c2fnas, chen2021cross, guan2021radgraph}. Graph Convolutional Networks (GCNs)~\cite{kipf2016semi, velickovic2017graph}, due to their capability in capturing complex node associations, have been widely applied in medical image analysis. In this context, representative works (e.g., CheXGCN ~\cite{chexgcn}) have pioneered the use of graph networks to learn label co-occurrence relationships in chest X-ray image classification~\cite{hou2021ratchet, zhang2020when}, demonstrating the potential of explicit topological structures in handling multi-label pathologies.

However, these methods have primarily focused on pure visual classification tasks and have not yet addressed how to transfer and utilize this structured topological knowledge to guide the complex Natural Language Generation (NLG) process~\cite{endo2021retrieval}. In this paper, inspired by the label co-occurrence learning paradigm, we adapt the TKI module for cross-modal report generation. We utilize the GCN mechanism within TKI to model the global statistical disease co-occurrence relationships within medical datasets and further employ its output topological representations as a "bridge" for cross-modal generation. Specifically, we integrate the explicit weights dynamically generated by the TKI module with the prompt-driven paradigm, utilizing them directly as the physical parameters of the main classification prompt generator via a "weight tying" mechanism. This design leverages the structural properties of graph networks in modeling complex complications and reduces the reliance of report generation models on external retrieval mechanisms, facilitating the injection of topological knowledge into the text generation phase in a parameterized manner.

\begin{figure*}[htbp]
    \centering
    \includegraphics[width=\textwidth]{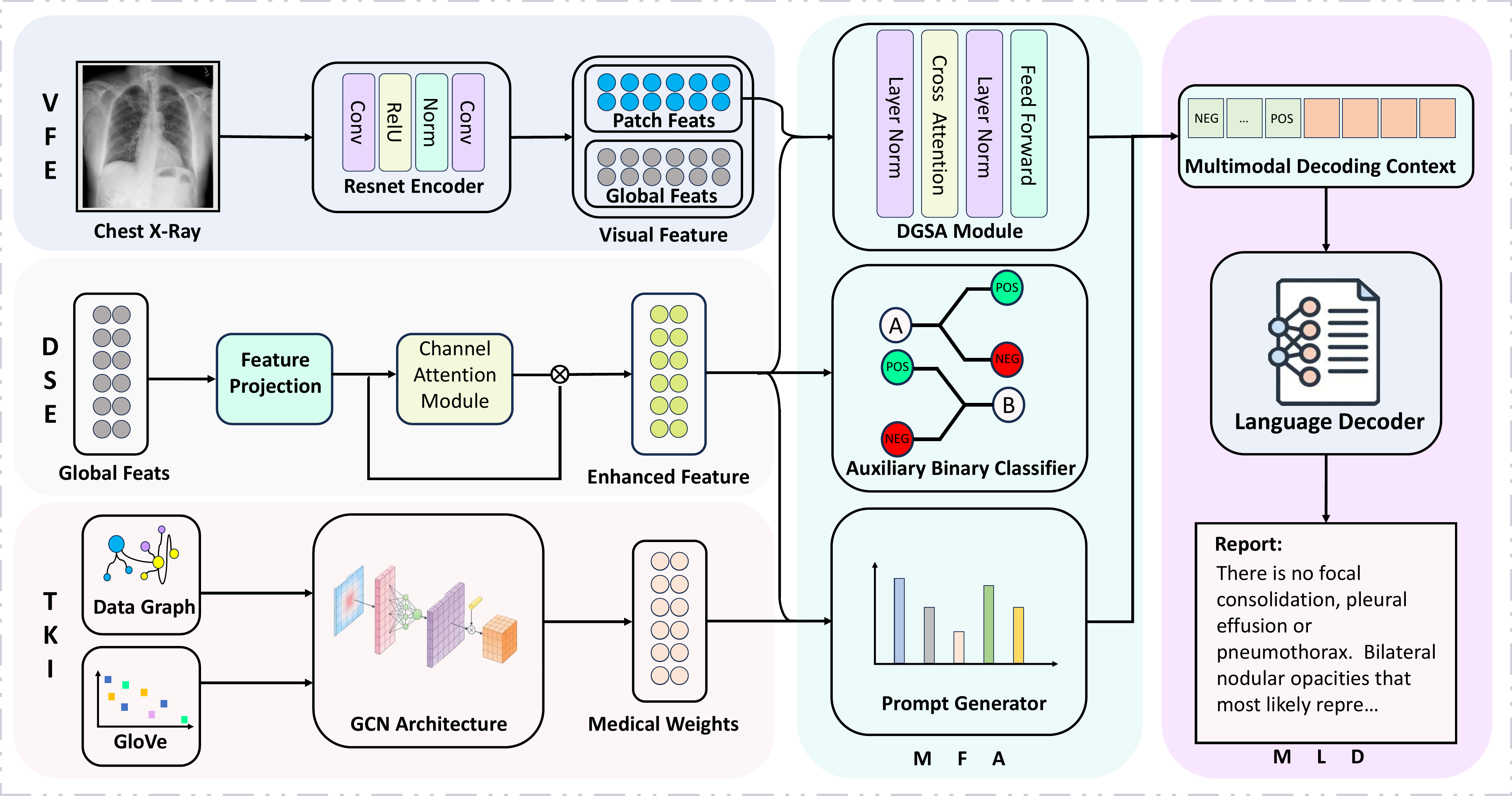} 
    \caption{The overall architecture of the proposed GDMRG framework. The system consists of five cohesive modules: Visual Feature Extraction (VFE), Diagnostic Semantic Enhancement (DSE), Topological Knowledge Internalization (TKI), Multimodal Feature Alignment (MFA), and Multimodal Language Decoding (MLD). Within the framework, an \textbf{Auxiliary Binary Classifier} is employed to optimize the intermediate diagnostic features. Subsequently, the MFA module acts as the core integration hub, bridging high-dimensional disease topologies with spatial visual grounding via the Diagnosis-Guided Spatial Attention (\textbf{DGSA}) mechanism.}
    \label{fig:overall_architecture}
\end{figure*}

\subsection{Long-Tailed Learning in Multi-Label Classification}
Multi-label medical data, such as chest X-rays, typically exhibit pronounced long-tailed distributions and class imbalances~\cite{zhang2023knowledge, cui2019class} (i.e., the majority of image regions represent healthy tissues, while positive lesions are relatively sparse). Recent explorations in Medical Report Generation (MRG) have sought to bridge the gap between classification and generation tasks by utilizing discrete diagnostic prompts~\cite{promptmrg}. These studies indicate that explicitly aligning intermediate classification outputs with the language decoder can enhance overall text generation. Despite the structural advantages of these frameworks, learning robust predictive representations from highly imbalanced pathological distributions remains an inherent challenge in the medical domain~\cite{cao2019learning}.

To address this inherent difficulty, this paper introduces a domain-adapted Truncated Asymmetric Loss (T-ASL) into the auxiliary branch of our dual-stream classification system. Building upon the foundational Asymmetric Loss~\cite{ridnik2021asymmetric}, which dynamically decouples and decays the gradient weights of abundant, easily identifiable negative samples, T-ASL further incorporates a bidirectional gradient truncation mechanism. Rather than executing explicit numerical interventions on the data distribution, this combined approach mitigates the dominance of healthy background samples and reduces the impact of medical label noise. Consequently, this mechanism encourages the network to focus its optimization more effectively on sparse, hard-to-classify positive targets. Coupled with the Optimal Threshold Selection (OTS) strategy employed during the inference phase, this approach improves the model's sensitivity to long-tailed positive lesions while maintaining its fidelity to the underlying clinical probability distribution.

\section{Methodology}

Figure \ref{fig:overall_architecture} shows the schematic illustration of the proposed Graph-Augmented Dual-Stream Medical Report Generation with Topological Internalization (GDMRG). To handle the inherent challenges of visual grounding and long-tailed diagnostic omissions in medical report generation, our framework is formulated as an end-to-end data flow.

As depicted in the architectural blueprint, the model processes multimodal information through three parallel foundational streams that ultimately converge. Specifically, the top stream (VFE) focuses on extracting localized patch features from the input radiological image and deriving aggregated global features via global average pooling. The middle stream (DSE) is responsible for distilling these aggregated global features into compact diagnostic semantics. Concurrently, the bottom stream (TKI) constructs a parametric topological graph based on statistical disease co-occurrences. These three independent streams converge at the Multimodal Feature Alignment (MFA) module, which acts as a routing hub to generate diagnostic prompts and execute diagnostic-driven visual recalibration. Finally, the Multimodal Language Decoding (MLD) module takes the aligned context to autoregressively generate the medical report.

\subsection{Visual Feature Extraction (VFE)}
First, a radiological image is input into a pre-trained ResNet-101 backbone to extract dense visual features. To encode more local information and preserve the fine-grained spatial structural integrity of the chest X-ray, we discard the global average pooling layer and the fully connected layer of the CNN. This process yields the localized patch features $\mathbf{V}_{patch} \in \mathbb{R}^{B \times N \times D}$, where $B$ is the batch size, $N=49$ denotes the number of localized spatial patches, and $D=2048$ is the channel dimension. Subsequently, a global average pooling operation is applied directly to these patch features to derive the aggregated global features $\mathbf{v}_{global} \in \mathbb{R}^{B \times D}$.

\subsection{Diagnostic Semantic Enhancement (DSE)}
Due to the subtle and morphologically ambiguous nature of medical lesions, directly utilizing the pooled global features $\mathbf{v}_{global}$ for downstream long-tailed classification often causes rare abnormalities to be submerged in background noise. To mitigate this issue, we introduce the DSE module to reduce visual redundancy and extract task-relevant high-frequency responses. 

Specifically, the global feature first undergoes a linear feature projection to map the visual semantics into a medical diagnostic subspace, generating the basis feature $\mathbf{v}_{proj}$. Next, this feature is fed into a bottleneck-style Channel Attention Module. This module employs a Squeeze-and-Excitation paradigm~\cite{hu2018squeeze}, explicitly modeling inter-channel dependencies to dynamically calculate the channel excitation weights $\mathbf{w}_c \in (0,1)^{D_x}$. The enhanced diagnostic feature $\mathbf{x}$ is then obtained via channel-wise scaling followed by a residual connection:
\begin{equation}
    \mathbf{x} = \mathbf{v}_{proj} + (\mathbf{v}_{proj} \otimes \mathbf{w}_c), \quad \mathbf{x} \in \mathbb{R}^{B \times D_x}
\end{equation}
This operation results in more compact and task-relevant condensed visual features, preparing a high signal-to-noise ratio foundation for the subsequent semantic alignment.

\subsection{Topological Knowledge Internalization (TKI)}
\begin{figure}[htbp]
    \centering
    \includegraphics[width=0.95\linewidth]{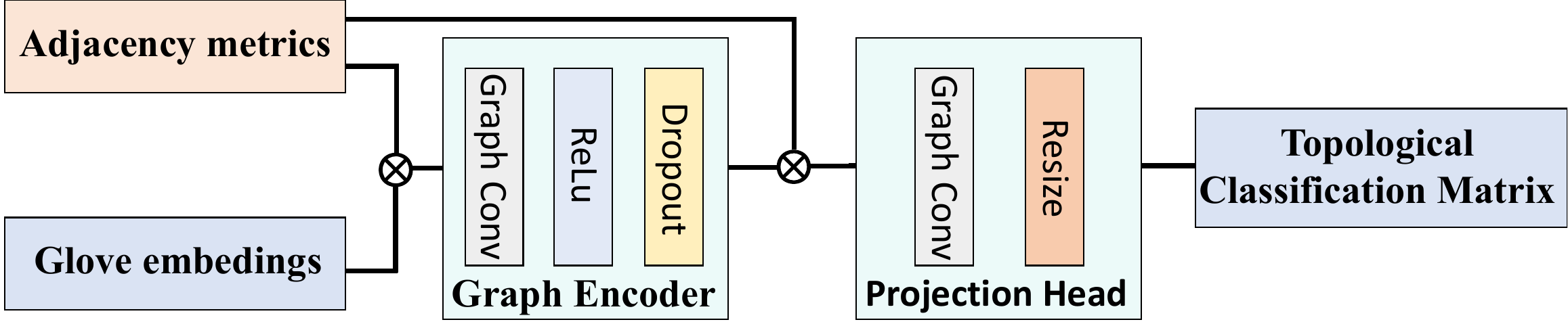}
    \caption{Detailed architecture of the proposed Topological Knowledge Internalization (TKI) module. To delineate the topological learning process, the network is conceptually divided into a Graph Encoder and a Projection Head. The Graph Encoder extracts and regularizes topological features from initial embeddings and adjacency metrics. The Projection Head subsequently expands these latent representations and applies a Resize operation to decouple the clinical states, ultimately generating the dynamic parameter space for the dual-stream classifier.}
    \label{fig:tki_flow}
\end{figure}

In clinical practice, abnormalities are rarely isolated; radiologists often rely on anatomical and complication priors during diagnosis. Relying solely on independent visual features may be insufficient when deciphering complex disease co-occurrences. To transform these medical priors into parameterized constraints, we introduce the Topological Knowledge Internalization (TKI) module, as illustrated in Fig.~\ref{fig:tki_flow}.

\textbf{Graph Construction and Normalization:} We define $N_{d}=18$ specific concept nodes to construct the graph, comprising 14 core chest abnormalities and 4 auxiliary anatomical attributes (the exact annotation protocol is detailed in Section 4.1). Let $\mathbf{H}^{(0)} \in \mathbb{R}^{N_{d} \times 300}$ represent the initial node embeddings. To construct the adjacency metrics, we calculate the raw co-occurrence frequency matrix $\mathbf{M}$ across the training set. To mitigate the bias caused by varying baseline disease frequencies, we apply geometric mean normalization:
$$
    \mathbf{M}'_{ij} = \frac{\mathbf{M}_{ij}}{\sqrt{\mathbf{M}_{ii} \mathbf{M}_{jj}}}
$$
Subsequently, we apply a topological sparsity threshold $\phi$ (empirically set to the 90th-percentile, retaining the top 10\% of edges) to $\mathbf{M}'_{ij}$ to binarize the edges, retaining only the principal topological connections. After adding identity self-loops ($\mathbf{\hat{A}} = \mathbf{A}_{thresh} + \mathbf{I}$), we compute the symmetric normalized adjacency matrix $\mathbf{\tilde{A}} = \mathbf{D}^{-1/2} \mathbf{\hat{A}} \mathbf{D}^{-1/2}$.

\textbf{Feature Propagation and Projection:} As depicted in Fig.~\ref{fig:tki_flow}, the derived adjacency metrics $\mathbf{\tilde{A}}$ and initial embeddings $\mathbf{H}^{(0)}$ are fed into a sequential architecture to learn the topological semantics. The network consists of a \textbf{Graph Encoder} and a \textbf{Projection Head}. In the Graph Encoder, the initial features are updated through a graph convolution operation ($\mathbf{\tilde{A}} \mathbf{H}^{(0)} \mathbf{W}_{gcn}^{(0)}$), followed by a non-linear ReLU activation and a Dropout operation to mitigate overfitting. This intermediate representation $\mathbf{H}^{(1)}$ is then routed through the Projection Head. Here, a second graph convolution expands the latent space, which subsequently undergoes a \textit{Resize} (state decoupling) operation to form 4 discrete sub-spaces. Formally, the propagation process is formulated as:
$$
    \mathbf{H}^{(1)} = \text{Dropout}\left(\text{ReLU}\left( \mathbf{\tilde{A}} \mathbf{H}^{(0)} \mathbf{W}_{gcn}^{(0)} \right)\right)
$$
$$
    \mathbf{W} = \text{Reshape}\left( \mathbf{\tilde{A}} \mathbf{H}^{(1)} \mathbf{W}_{gcn}^{(1)} \right)
$$
where the channel dimensions transition from 300 to 1024, and finally to a flattened state space of $C \times D_{h} = 2048$. The output matrix is subsequently reshaped to generate the Topological Classification Matrix $\mathbf{W} \in \mathbb{R}^{(N_{d} \times C) \times D_{h}}$, where $C=4$ represents the discrete clinical states ([POS], [NEG], [BLA], [UNC]) and $D_{h}=512$ corresponds to the decoupled semantic vectors. Serving as the dynamic parameter space for the main classifier, this pathway encourages the final predictions to align with established medical priors.

\subsection{Multimodal Feature Alignment (MFA)} To enable the text decoder to focus on specific pathological regions accurately, the T-ASL-refined diagnostic feature $\mathbf{f}_{cond}$ serves as a Query ($\mathbf{Q}$). It interacts with the dense visual features $\mathbf{V}_{spatial}$ (acting as Key $\mathbf{K}$ and Value $\mathbf{V}$) via a multi-head Diagnosis-Guided Spatial Attention (DGSA)~\cite{vaswani2017attention}. Following standard transformer blocks (comprising layer normalization and feed-forward networks with residual connections), this process yields a pathology-focused spatial representation $\mathbf{V}_{attn\_out} \in \mathbb{R}^{B \times N \times D}$.

While $\mathbf{V}_{attn\_out}$ provides concentrated visual evidence for specific lesions, excessive reliance on it may compromise the broader anatomical context. Rather than utilizing a complex independent fusion module, we introduce a lightweight dynamic gating scalar $g \in (0, 1)$ to fuse the global visual context with the localized pathological focus. Specifically, $g$ is derived by applying global average pooling to both $\mathbf{V}_{spatial}$ and $\mathbf{V}_{attn\_out}$, concatenating them, and passing the result through a Sigmoid-activated Multilayer Perceptron. The final fused spatial feature is then computed via a convex combination: $\mathbf{V}_{fused} = (1 - g) \cdot \mathbf{V}_{spatial} + g \cdot \mathbf{V}_{attn\_out}$. This adaptive scalar functions as a soft switch, dynamically routing spatial information prior to language decoding.

\subsection{Multimodal Language Decoding (MLD)}

Finally, an autoregressive language decoder (initialized with pre-trained BERT weights) processes the discrete diagnostic prompts as prefix inputs. It leverages the fused spatial context $\mathbf{V}_{fused}$ to generate clinically accurate reports. The text generation process is optimized using the Negative Log-Likelihood loss ($\mathcal{L}_{NLL}$):
\begin{equation}
    \mathcal{L}_{NLL} = - \sum_{t=1}^{T} \log P(w_t | w_{<t}, \mathbf{V}_{fused}, \text{prompt})
\end{equation}

The entire GDMRG framework is trained end-to-end, and the overall objective function is formulated as:
\begin{equation}
    \mathcal{L}_{total} = \lambda_{cls} (\mathcal{L}_{CE} + \mathcal{L}_{T-ASL}) + \lambda_{lm} \mathcal{L}_{NLL}
\end{equation}
where $\lambda_{cls}$ and $\lambda_{lm}$ are hyperparameters empirically tuned to balance the contributions of the classification and text decoding tasks, respectively.

\section{Experiments}

\subsection{Datasets and Annotation Protocol}
We evaluate the proposed GDMRG framework on two widely used public medical imaging datasets: MIMIC-CXR~\cite{johnson2019mimic} and IU X-Ray~\cite{demner2016preparing}. 

To bridge the gap between free-text reports and structured diagnostic supervision, we utilize an extended 18-dimensional annotation protocol~\cite{promptmrg}. Alongside the 14 core chest abnormalities derived via the standard CheXpert labeler~\cite{irvin2019chexpert}, we incorporate 4 anatomical attributes (\textit{Aorta}, \textit{Bone}, \textit{Hemidiaphragm}, and \textit{Lung Volume}) to provide more granular clinical context. Adopting this unified protocol ensures direct comparability with recent state-of-the-art methods.

Based on this 18-class annotation protocol, the datasets are processed as follows:
\begin{itemize}
    \item \textbf{MIMIC-CXR}: This represents the largest publicly available chest X-ray dataset to date. We adhere to the dataset splits and preprocessing procedures established in prior MRG literature~\cite{chen2020generating}. The 18-dimensional labels are utilized to provide supervision for our dual-stream classifier and to define the topological nodes in the TKI module.
    
    \item \textbf{IU X-Ray}: A smaller-scale dataset comprising 7,470 images and 3,955 reports. In clinical report generation, evaluation on small-scale splits often presents challenges for disease-specific assessments, as rare abnormalities may lack sufficient positive samples to yield statistically meaningful results. Recognizing this limitation in traditional intra-dataset splits~\cite{chen2020generating}, we follow the zero-shot evaluation paradigm proposed in recent studies~\cite{promptmrg}. Specifically, we utilize the model trained on the MIMIC-CXR training set to directly perform evaluation on the IU X-Ray dataset. This cross-domain setup allows for a more robust assessment of the model's generalization capabilities and its diagnostic reliability across all 18 pathological categories.
\end{itemize}

\subsection{Evaluation Metrics}
We employ two sets of evaluation metrics: Natural Language Generation (NLG) and Clinical Efficacy (CE). 
The NLG metrics include BLEU-1 through 4~\cite{papineni2002bleu}, METEOR~\cite{banerjee2005meteor}, and ROUGE-L~\cite{lin2004rouge}. 
The CE metrics utilize CheXpert to evaluate the model's Precision, Recall, and F1-score across the 14 core diseases. To account for variations in CE calculation protocols across current literature and ensure a comprehensive evaluation, we report the test results based on both instance-level macro-averaging and global micro-averaging in our main experiments.

\begin{table*}[htbp]
\centering
\caption{Comprehensive comparison with SOTA MRG methods on MIMIC-CXR and IU X-Ray datasets. '-' denotes metrics that are unavailable or omitted.}
\label{tab:main_results}

\vspace{0.1cm}
\textbf{(a) Results on the MIMIC-CXR dataset (CE and NLG Metrics)} \\
\vspace{0.1cm}
\begin{tabular*}{0.85\textwidth}{@{\extracolsep{\fill}} ll ccc cccc @{}}
\toprule
\multirow{2}{*}{\textbf{Model}} & \multirow{2}{*}{\textbf{Year}} & \multicolumn{3}{c}{\textbf{CE Metrics}} & \multicolumn{4}{c}{\textbf{NLG Metrics}} \\
\cmidrule(lr){3-5} \cmidrule(lr){6-9}
 & & Precision & Recall & F1 & BL-1 & BL-4 & METEOR & ROUGE-L \\
\midrule
R2Gen~\cite{chen2020generating} & 2020 & 0.333 & 0.273 & 0.276 & 0.353 & 0.103 & 0.142 & 0.277 \\
M2TR~\cite{m2tr} & 2021 & 0.240 & 0.428 & 0.308 & 0.378 & 0.107 & 0.145 & 0.272 \\
CA~\cite{liu2021contrastive} & 2021 & 0.352 & 0.298 & 0.303 & 0.350 & 0.109 & 0.151 & 0.283 \\
CVT2Dis~\cite{cvt2dis} & 2022 & 0.356 & 0.412 & 0.384 & 0.392 & \textbf{0.124} & 0.153 & 0.285 \\
M2KT~\cite{m2kt} & 2023 & 0.420 & 0.339 & 0.352 & 0.386 & 0.111 & - & 0.274 \\
METrans~\cite{metrans} & 2023 & 0.364 & 0.309 & 0.311 & 0.386 & \textbf{0.124} & 0.152 & \textbf{0.291} \\
DCL~\cite{dcl} & 2023 & 0.471 & 0.352 & 0.373 & - & 0.109 & 0.150 & 0.284 \\
PromptMRG$^{\dagger}$~\cite{promptmrg} & 2024 & 0.503 & 0.493 & 0.469 & 0.391 & 0.105 & 0.152 & 0.266 \\
CAMANet~\cite{camanet} & 2024 & 0.483 & 0.323 & 0.387 & 0.374 & 0.112 & 0.145 & 0.279 \\
Teaser~\cite{teaser} & 2025 & \textbf{0.534} & 0.518 & 0.526 & \textbf{0.423} & 0.113 & \textbf{0.170} & 0.287 \\
CMCRL~\cite{cmcrl} & 2025 & 0.489 & 0.340 & 0.401 & 0.400 & 0.119 & 0.150 & 0.280 \\
\midrule
\textbf{Ours} & - & 0.519 & \textbf{0.610} & \textbf{0.561} & 0.406 & 0.114 & 0.157 & 0.272 \\
\bottomrule
\end{tabular*}

\vspace{0.4cm} 

\textbf{(b) Results on the IU X-Ray dataset (Zero-Shot NLG Metrics)} \\
\vspace{0.1cm}
\begin{tabular*}{0.85\textwidth}{@{\extracolsep{\fill}} ll cccccc @{}}
\toprule
\multirow{2}{*}{\textbf{Model}} & \multirow{2}{*}{\textbf{Year}} & \multicolumn{6}{c}{\textbf{NLG Metrics}} \\
\cmidrule(lr){3-8} 
 & & BL-1 & BL-2 & BL-3 & BL-4 & METEOR & ROUGE-L \\
\midrule
R2Gen$^{\dagger}$~\cite{chen2020generating} & 2020 & 0.335 & 0.180 & 0.103 & 0.063 & 0.133 & 0.257 \\
CVT2Dis$^{\ddagger}$~\cite{cvt2dis} & 2022 & 0.383 & - & - & 0.082 & 0.147 & 0.277 \\
M2KT$^{\ddagger}$~\cite{m2kt} & 2023 & 0.371 & - & - & 0.078 & 0.153 & 0.261 \\
DCL$^{\ddagger}$~\cite{dcl} & 2023 & 0.354 & - & - & 0.074 & 0.152 & 0.267 \\
RGRG$^{\ddagger}$~\cite{rgrg} & 2023 & 0.266 & - & - & 0.063 & 0.146 & 0.180 \\
PromptMRG$^{\ddagger}$~\cite{promptmrg} & 2024 & 0.401 & - & - & 0.098 & \textbf{0.160} & 0.281 \\
CAMANet$^{\dagger}$ & 2024 & 0.359 & 0.198 & 0.115 & 0.071 & 0.139 & 0.272 \\
CMCRL$^{\dagger}$ & 2025 & 0.367 & 0.200 & 0.112 & 0.066 & 0.134 & 0.255 \\
\midrule
\textbf{Ours} & - & \textbf{0.414} & \textbf{0.248} & \textbf{0.161} & \textbf{0.109} & 0.159 & \textbf{0.313} \\
\bottomrule
\multicolumn{8}{l}{\footnotesize $^{\dagger}$ indicates performance evaluated by us.} \\
\multicolumn{8}{l}{\footnotesize $^{\ddagger}$ indicates performance evaluated and reported by PromptMRG.} \\
\end{tabular*}
\end{table*}

\subsection{Implementation Details}

\textbf{Hardware Setup and Two-stage Training Paradigm:} All experiments were implemented using PyTorch and conducted on a single NVIDIA GeForce RTX 4090 GPU. We utilized a pre-trained ResNet as the visual backbone, GloVe embeddings for TKI node features, and a pre-trained BERT~\cite{devlin2018bert} as the text decoder. To ensure stable convergence, we employed a two-stage training paradigm: first, a warm-up training exclusively on the dual-stream classifier system; second, an end-to-end joint training of the complete "vision-classification-text" pipeline. 

\textbf{Optimization and Inference Setup:} The model was optimized using AdamW. To prevent catastrophic forgetting while accelerating the convergence of newly initialized modules, we implemented a hierarchical learning rate strategy, complemented by a ReduceLROnPlateau scheduling. Furthermore, to stabilize the dynamic spatial fusion, a batch-wise curriculum learning strategy was introduced during the initial joint training phase. During inference, we employed beam search and determined classification thresholds via grid search for discrete class assignment. Dataset-specific preprocessing was applied: resizing for MIMIC-CXR and center crop for IU X-Ray to 224 x 224 to handle distinct dataset characteristics. The complete source code, pre-trained network weights, and exact training schedules will be made publicly available upon formal acceptance.

\subsection{Main Results}

To comprehensively evaluate the generation quality and clinical efficacy of our proposed GDMRG framework, we compare it against a wide range of state-of-the-art (SOTA) MRG methods on both the MIMIC-CXR and IU X-Ray datasets. The quantitative results are summarized in Table \ref{tab:main_results}.

\textbf{1) Clinical Efficacy (CE) Metrics:} 
The primary objective of automated medical report generation is to provide precise diagnostic evidence to assist radiological professionals. Consequently, CE metrics, which evaluate the accuracy of abnormality classification, serve as crucial indicators of a model's clinical utility. As demonstrated in \textbf{Table \ref{tab:main_results}(a)}, GDMRG achieves competitive overall clinical efficacy on the MIMIC-CXR dataset, with a Precision of 0.519. Specifically, our model records a Recall of 0.610 and an F1-score of 0.561.

Notably, regarding Recall, GDMRG achieves an absolute improvement of 9.2\% over the recent baseline, Teaser (2025). This enhancement indicates a reduction in missed diagnoses (false negatives) for rare or subtle lesions. We attribute this success to the synergy of our Topological Knowledge Internalization (TKI) module and the domain-adapted Truncated Asymmetric Loss (T-ASL) optimization strategy. By explicitly modeling global disease co-occurrence and dynamically down-weighting abundant healthy background samples while truncating noisy gradients from ambiguous outliers, the model addresses the long-tailed distribution challenge, thereby improving the detection of sparse anomalies.

\textbf{2) Natural Language Generation (NLG) Metrics and Clinical Trade-offs:}
In addition to diagnostic accuracy, we evaluate the text generation quality using standard NLG metrics. On the MIMIC-CXR dataset, we note that our method yields lower performance across standard N-gram metrics, trailing the scores reported by text-centric baselines like CVT2Dis and METrans on metrics such as BLEU-4. 

We posit that this phenomenon reflects an inherent and well-documented architectural trade-off within the MRG domain. Standard NLG metrics primarily evaluate the N-gram consistency between predictions and references. Consequently, generating high-frequency, standard anatomical descriptions (which constitute the vast majority of normal training corpora) naturally yields higher BLEU scores. Conversely, GDMRG is heavily regularized to prioritize sparse, positive pathological findings. Inserting these critical, highly specific diagnostic descriptions inevitably disrupts the standard, frequent N-gram sequences typical of normal reports. Considering the substantial 3.5\% absolute improvement in the comprehensive CE F1-score over the highest baseline, we suggest that this compromise in N-gram syntactic matching is an acceptable and clinically relevant trade-off, underscoring our model's commitment to diagnostic faithfulness over surface-level n-gram matching.

\textbf{3) Zero-Shot Generalization on IU X-Ray:}
Furthermore, to assess the cross-domain robustness of our framework, we conducted zero-shot testing on the IU X-Ray dataset (\textbf{see Table \ref{tab:main_results}(b)}). Following the evaluation protocols of recent SOTA works (e.g., PromptMRG, CMCRL), we omit the CE metrics for IU X-Ray due to the domain shift associated with the CheXpert labeler, which was specifically engineered for the MIMIC-CXR linguistic space. 

Under this zero-shot setting without any domain-specific fine-tuning, GDMRG demonstrates improved performance across the evaluated NLG metrics compared to the referenced baselines. Specifically, it improves upon PromptMRG by 1.3\% in BLEU-1, 1.1\% in BLEU-4, and 3.2\% in ROUGE-L. This consistent performance supports the efficacy of our DGSA module. Even when confronting a domain shift, the DGSA module utilizes diagnostic queries to re-calibrate spatial visual features, assisting the decoder in generating coherent medical narratives.

\subsection{Ablation Study}
To evaluate the specific contribution of each proposed architectural module and optimization strategy within the GDMRG framework, we conducted a comprehensive ablation study on the MIMIC-CXR dataset. The quantitative progression is detailed in Table \ref{tab:ablation_components}.

\subsubsection{Effectiveness of Core Components and Threshold Calibration} \hfill \\
\hspace*{\parindent}\textbf{1) The Role of Optimal Threshold Selection (OTS):} 
Table \ref{tab:ablation_components} illustrates the performance shift across model variants when the Optimal Threshold Selection (OTS) strategy is applied. Using standard thresholding (Row 1), the BASE model exhibits a metric imbalance, yielding a Precision of 0.606 and a lower Recall of 0.472, indicating that sparse positive lesions are frequently missed. By introducing OTS to dynamically calibrate the decision boundary (Row 2), the model trades a portion of Precision for an increased Recall (0.592), raising the F1-score to 0.537 and BLEU-1 to 0.402. This suggests that adaptive thresholding is an effective strategy for addressing imbalanced medical distributions.

\textbf{2) Impact of Independent Modules:} 
Building upon the OTS-calibrated baseline, we evaluate the architectural extensions. The integration of the DSE and DGSA modules (Row 4) yields consistent improvements, raising the F1-score to 0.546. Alternatively, when integrating the TKI module (Row 6), the diagnostic behavior shifts toward a higher sensitivity. Guided by explicit pathological co-occurrence priors, this variant achieves the highest overall Recall rate of \textbf{0.616}. This indicates that topological knowledge assists in identifying sparse and complex concurrent lesions that might be overlooked by purely visual features. However, relying heavily on global disease priors introduces a slight tendency for over-association, leading to a marginal drop in BLEU-1 (0.399).

\textbf{3) Synergy in the Full Architecture:} 
The complete GDMRG framework addresses this trade-off by combining TKI with the DGSA module. Without OTS (Row 7), the full architecture favors diagnostic certainty, resulting in the dataset's highest Precision (\textbf{0.618}) but demonstrating limited generalization to long-tailed positive samples. However, when calibrated with OTS (Row 8), the DGSA module acts as a visual-grounding filter, verifying the topological hypotheses generated by TKI against the actual visual evidence. Consequently, the full model balances the metrics, yielding a Precision of 0.519 while maintaining a Recall of 0.610. This synergistic mechanism achieves the highest overall F1-score of \textbf{0.561} and the highest BLEU-1 score of \textbf{0.406}, demonstrating the effectiveness and structural rationale of each component in the proposed architecture.

\begin{table}[htbp]
\centering
\caption{Ablation study of different network components. \textbf{OTS}: Optimal Threshold Selection; \textbf{DSE}: Diagnostic Semantic Enhancement; \textbf{TKI}: Topological Knowledge Internalization; \textbf{DGSA}: Diagnosis-Guided Spatial Attention.}
\label{tab:ablation_components}
\setlength{\tabcolsep}{4pt}
\resizebox{\columnwidth}{!}{
    \begin{tabular}{lcccc}
    \toprule
    \textbf{Models} & Precision & Recall & F1 & BLEU-1 \\
    \midrule
    BASE & 0.606 & 0.472 & 0.531 & 0.388 \\
    BASE + OTS & 0.492 & 0.592 & 0.537 & 0.402 \\
    BASE + DSE + DGSA & 0.606 & 0.475 & 0.532 & 0.383 \\
    BASE + OTS + DSE + DGSA & 0.498 & 0.605 & 0.546 & 0.404 \\
    BASE + DSE + TKI & 0.607 & 0.497 & 0.546 & 0.396 \\
    BASE + OTS + DSE + TKI & 0.503 & \textbf{0.616} & 0.554 & 0.399 \\
    BASE + DSE + TKI + DGSA & \textbf{0.618} & 0.471 & 0.534 & 0.383 \\
    BASE + OTS + DSE + TKI + DGSA & 0.519 & 0.610 & \textbf{0.561} & \textbf{0.406} \\
    \bottomrule
    \end{tabular}
}
\end{table}

\textbf{4) Supplementary Sample-Level Evaluation:}
To facilitate direct comparison with alternative benchmarking protocols (e.g., PromptMRG), we provide an additional ablation study based on sample-level (instance-level) clinical efficacy metrics. Due to space constraints and to maintain consistent baseline comparisons with micro-averaged literature in the main text, these supplementary results are detailed in the \textbf{Supplementary Material} (Table S1). Under this sample-level protocol, GDMRG maintains its relative performance gains, supporting the consistency of the proposed dual-stream framework.

\begin{table}[htbp]
\centering
\caption{Ablation study of different loss functions for the dual-stream classifier.}
\label{tab:ablation_loss}
\setlength{\tabcolsep}{5pt}
\resizebox{\columnwidth}{!}{
    \begin{tabular}{cccccc}
    \toprule
    \textbf{Main Loss} & \textbf{Binary Loss} & Precision & Recall & F1 & BLEU-1 \\
    \midrule
    CE & None & 0.502 & 0.597 & 0.545 & 0.402 \\
    CE & MBCE & 0.510 & 0.592 & 0.548 & 0.401 \\
    WFL & MBCE & 0.494 & 0.590 & 0.538 & \textbf{0.412} \\
    CE & ASL & 0.510 & 0.604 & 0.553 & 0.403 \\
    CE & T-ASL & \textbf{0.519} & \textbf{0.610} & \textbf{0.561} & 0.404 \\
    \bottomrule
    \end{tabular}
}
\end{table}

\subsubsection{Effectiveness of Optimization Strategies}
Addressing the highly imbalanced distribution of medical data requires robust gradient optimization. Table \ref{tab:ablation_loss} demonstrates the performance variations when adopting different loss functions within our dual-stream architecture. 

When the auxiliary branch introduces Masked Binary Cross Entropy (MBCE), the model achieves a higher Precision (0.510), but the Recall slightly decreases (0.592). This indicates that while MBCE filters out some false positives, it remains constrained in uncovering sparse positive lesions. Furthermore, if the main branch employs Weighted Focal Loss (WFL), the explicit reweighting of the probability distribution negatively impacts the overall diagnostic balance, causing the CE F1-score to decline to 0.538.

When the auxiliary branch utilizes the standard Asymmetric Loss (ASL), the model dynamically decouples and suppresses the gradients of abundant simple negative samples. This mechanism encourages the network to learn sparse positive features without substantially altering the main classification space, leading to notable improvements (Recall 0.604, F1 0.553). However, standard ASL remains susceptible to large penalty gradients induced by ambiguous or mislabeled hard samples common in clinical datasets. Ultimately, the application of our proposed Truncated Asymmetric Loss (T-ASL) addresses this limitation. By introducing a bidirectional clipping mechanism to mitigate the impact of label noise during optimization, T-ASL further increases the Recall to 0.610 and achieves the highest comprehensive F1-score of 0.561. This demonstrates that domain-adapted loss truncation is an effective strategy for mitigating the medical multi-label long-tail problem.

\subsection{Discussion: Macro to Micro Evidence of Topological Internalization}

To systematically evaluate the efficacy of the Topological Knowledge Internalization (TKI) module and address potential concerns regarding graph over-smoothing or multi-task interference, we present a comprehensive analysis spanning macroscopic parametric behaviors, quantitative sub-population evaluations, and microscopic visual grounding.

\subsubsection{Topological Sensitivity and Node Dimensionality}
To further evaluate the rationale behind the TKI module's design, we conducted an independent sensitivity analysis on node dimensionality and graph construction strategies, as detailed in Table \ref{tab:ablation_topology}. Notably, when utilizing only the 14 core abnormalities, our proposed TKI strategy does not substantially outperform the vanilla statistical co-occurrence baseline (Vanilla GCN). However, with the integration of the 4 auxiliary anatomical attributes (18-Node configuration), the TKI module achieves its highest performance (F1-score 0.561). This suggests that these auxiliary attributes act as important topological hubs, providing structural constraints that help mitigate the ambiguity of complex concurrent diseases. This analysis indicates that the synergy between the 18-node structure and our geometric normalization is highly beneficial for effective topological internalization.

\begin{table}[htbp]
\centering
\caption{Ablation study on node dimensionality and topological graph construction strategies. \textbf{Vanilla GCN} utilizes the raw statistical co-occurrence matrix widely adopted in previous works (e.g., CheXGCN). In contrast, \textbf{TKI} employs our proposed geometrically normalized and thresholded adjacency metrics. The \textbf{18-Node} configuration incorporates 4 additional anatomical attributes as crucial topological hubs.}
\label{tab:ablation_topology}
\setlength{\tabcolsep}{6pt}
\resizebox{\columnwidth}{!}{
    \begin{tabular}{lcccc}
    \toprule
    \textbf{Topological Configuration} & Precision & Recall & F1 & BLEU-1 \\
    \midrule
    14-Node Base (w/o Graph) & 0.510 & 0.590 & 0.547 & 0.399 \\
    14-Node + Vanilla GCN & 0.494 & \textbf{0.619} & 0.549 & 0.394 \\
    14-Node + TKI & 0.475 & 0.578 & 0.522 & 0.399 \\
    \midrule
    18-Node Base (w/o Graph) & 0.494 & 0.614 & 0.547 & \textbf{0.406} \\
    18-Node + Vanilla GCN & 0.497 & 0.600 & 0.544 & 0.390 \\
    18-Node + TKI (Ours) & \textbf{0.519} & 0.610 & \textbf{0.561} & \textbf{0.406} \\
    \bottomrule
    \end{tabular}
}
\end{table}

\subsubsection{Macroscopic Visualization of Topological Internalization}
A common challenge in applying Graph Convolutional Networks (GCNs) to dense classification tasks is the risk of "over-smoothing," where node representations collapse into indistinguishable states, thereby negating the topological structure. To verify whether the TKI module genuinely internalized the clinical priors, we visualize both the data-driven prior co-occurrence frequency matrix (derived from the training set) and the posterior parameterized weight matrix ($\mathbf{W}$) of the main classifier. 

\begin{figure}[htbp]
    \centering
    \includegraphics[width=\linewidth]{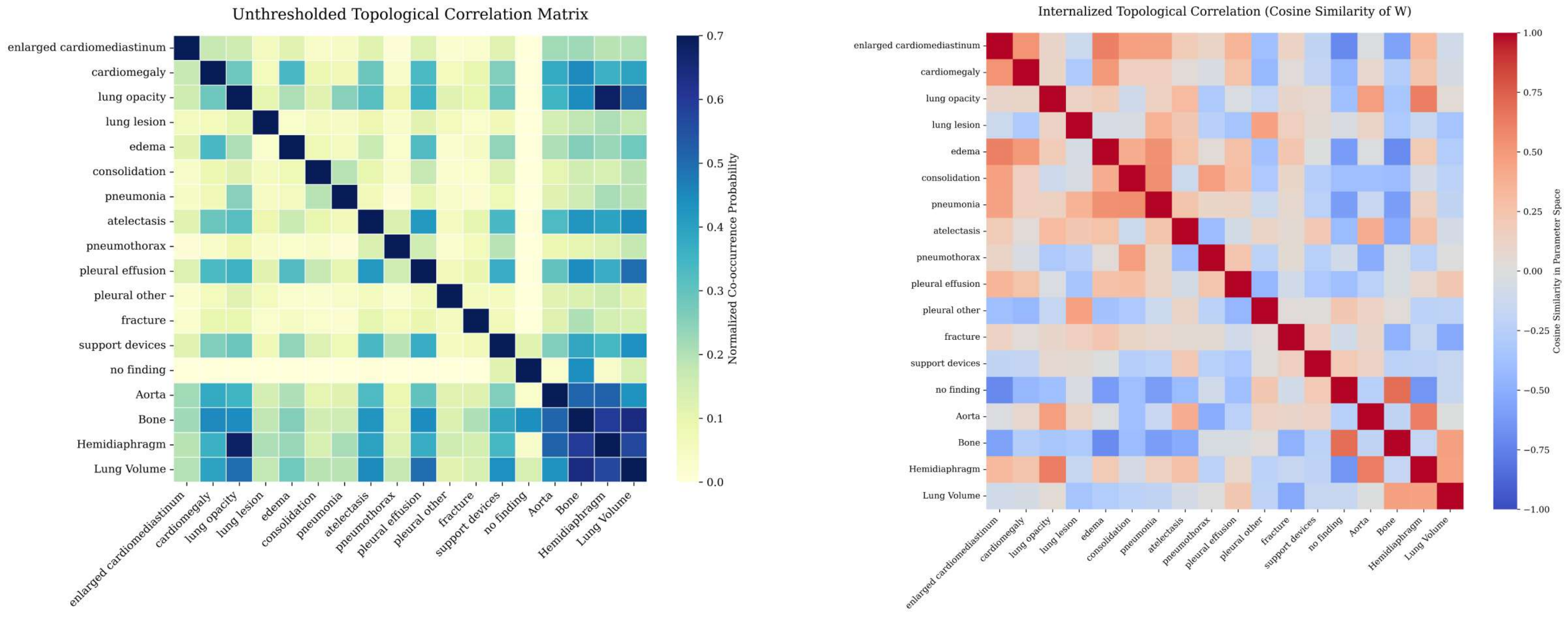} 
    \caption{(a) The 18-dimensional prior co-occurrence probability heatmap. (b) The mean-centered cosine similarity of the GCN-derived classifier weight matrix ($\mathbf{W}$). The alignment demonstrates successful topological internalization without over-smoothing.}
    \label{fig:topology_heatmap}
\end{figure}

As illustrated in Fig. \ref{fig:topology_heatmap}, we compute the Mean-Centered Cosine Similarity of the dynamically generated weight vectors in $\mathbf{W}$. The resulting heatmap demonstrates structural preservation rather than smoothing. Diseases with true clinical comorbidities (e.g., \textit{Cardiomegaly}, \textit{Pleural Effusion}, and \textit{Atelectasis}) exhibit strong positive cosine similarities, indicating synergistic activation within the continuous parameter manifold. Conversely, mutually exclusive or unrelated findings (e.g., \textit{Cardiomegaly} and \textit{Pneumothorax}) are naturally pushed into orthogonal or mutually inhibitory sub-spaces (represented by cool colors). Furthermore, the auxiliary anatomical tags act as essential topological hubs, validating that the TKI module successfully transforms discrete statistical priors into an explicit, physiologically sound parametric constraint.

\subsubsection{Performance under Complex Complications Subset}
Standard metrics on the entire dataset often dilute a model's performance on complex, concurrent lesions due to the high prevalence of healthy or single-disease samples. To isolate the specific contribution of the TKI module, we extracted a "Complex Subset" from the test set, exclusively comprising cases with two or more concurrent chest abnormalities (Complications $\ge 2$).

\begin{figure}[htbp]
    \centering
    \includegraphics[width=\linewidth]{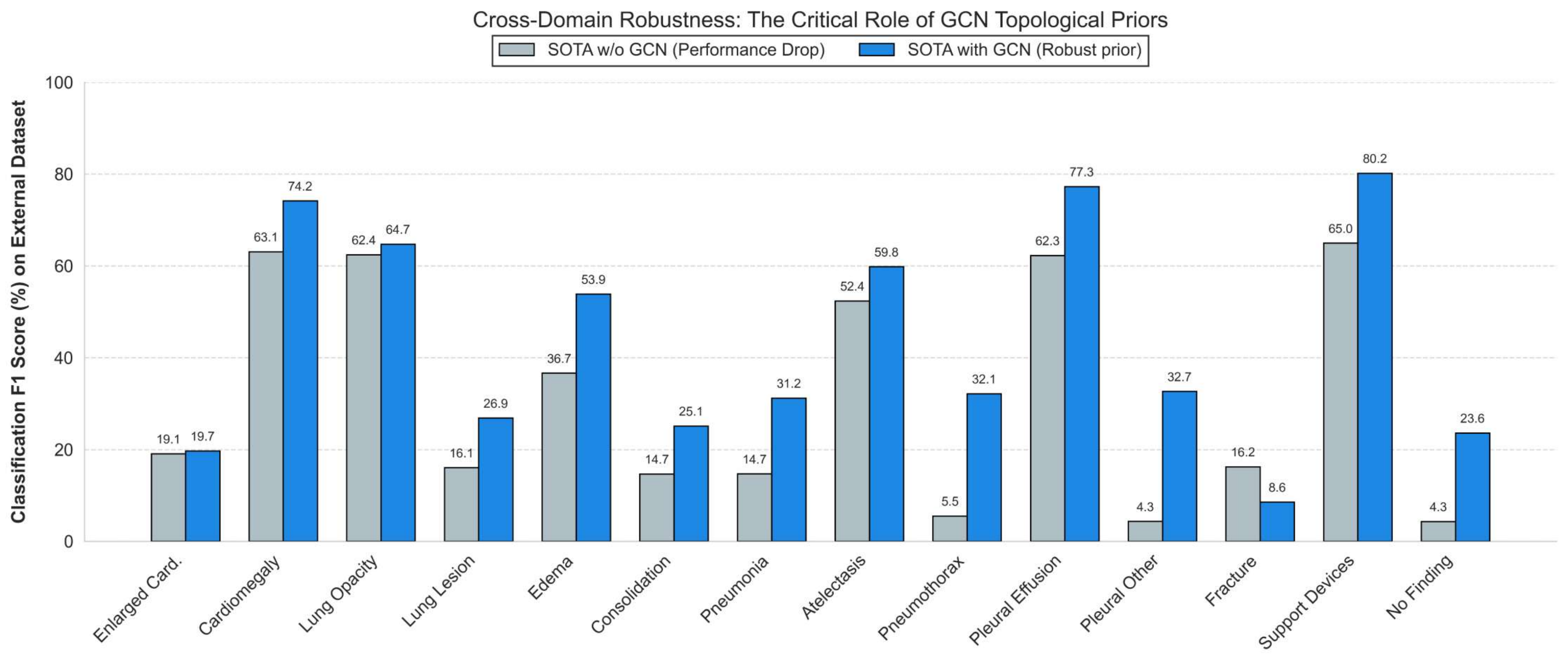} 
    \caption{Comparison of absolute F1 scores on the extracted Complex Subset (Complications $\ge 2$). The inclusion of TKI helps mitigate task interference for the majority of concurrent diseases, particularly for complex pleuropulmonary conditions.}
    \label{fig:complex_subset}
\end{figure}

As shown in Fig. \ref{fig:complex_subset}, we compare the absolute classification F1 scores of the intermediate classifier. In this challenging scenario, the Baseline model (without TKI) faces difficulties in balancing competing gradients from multiple concurrent labels. However, the introduction of explicit topological constraints acts as an endogenous regularizer. The chart reveals performance improvements across most long-tailed conditions, with the most substantial absolute gains observed in \textit{Pneumothorax}, \textit{Pleural Other}, and \textit{Pleural Effusion}. We also note a performance drop in isolated physical findings such as \textit{Fracture}, suggesting an inherent trade-off when applying global disease co-occurrence priors to highly localized, physically distinct injuries. Overall, this evaluation indicates that the TKI module helps stabilize multi-task learning under complex clinical presentations.

\subsubsection{Qualitative Analysis: From Linguistic Coherence to Visual Grounding}
To evaluate the efficacy of the Topological Knowledge Internalization (TKI) module, we conduct a dual-level qualitative analysis: a multi-sample textual comparison to demonstrate clinical relevance, followed by a fine-grained visual grounding inspection.

\textbf{1) Multi-Sample Linguistic Coherence and Hallucination Suppression:} 
Due to space constraints, the extended multi-sample textual comparison between our baseline (GDMRG w/o TKI) and the complete GDMRG across diverse clinical scenarios is detailed in the \textbf{Supplementary Material}. Notably, in the absence of explicit complication priors, the baseline often struggles with visually ambiguous entities, leading to clinical hallucinations (e.g., unsupported orthopaedic findings). Conversely, guided by the learned topological graph, GDMRG effectively mitigates these isolated hallucinations and captures complex clinical triads (e.g., concurrent cardiomegaly, pleural effusions, and atelectasis), yielding clinically consistent narratives.

\begin{figure}[htbp]
    \centering
    \includegraphics[width=\linewidth]{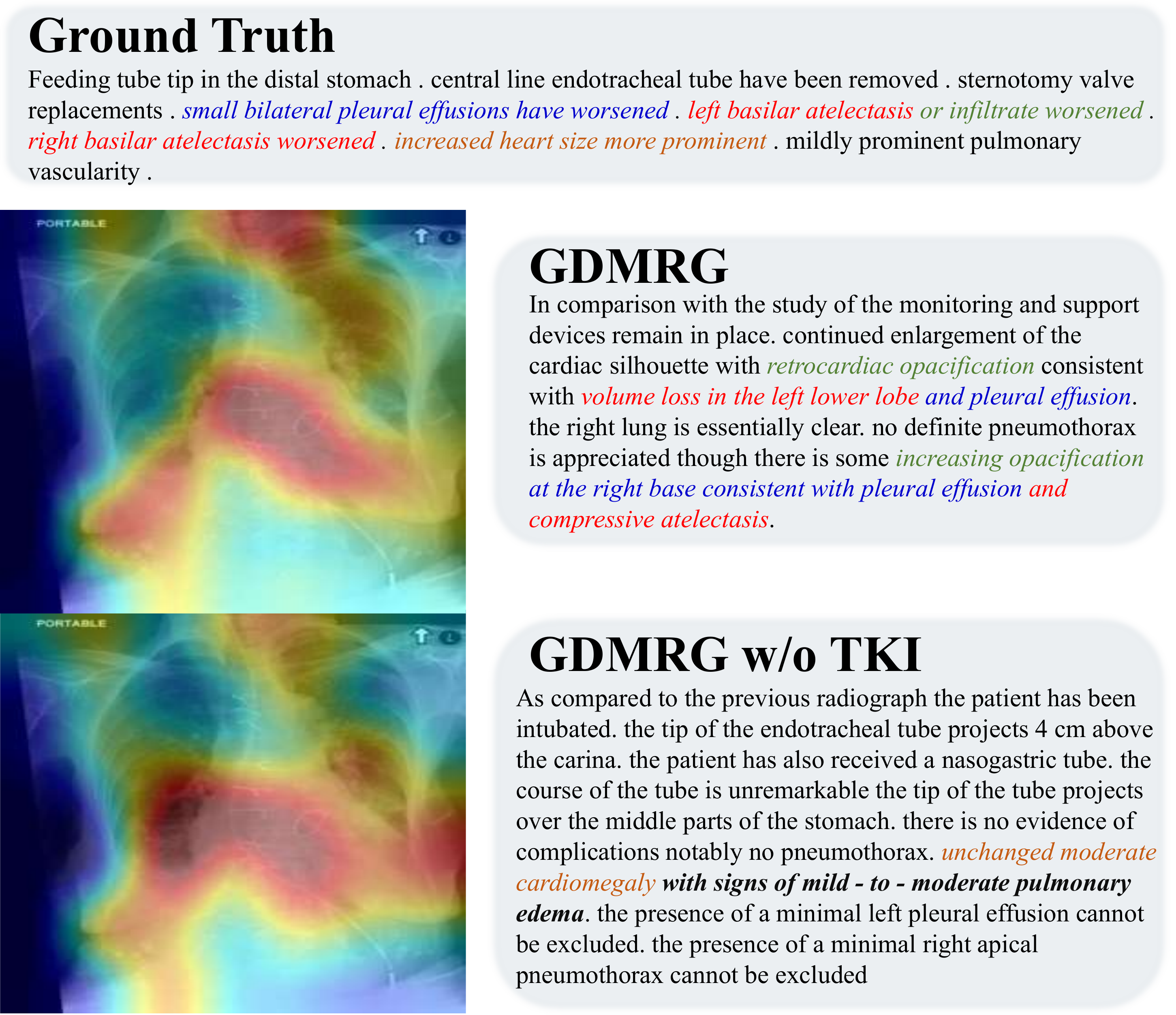} 
    \caption{\small Qualitative comparison of visual grounding and text generation. While the baseline model (w/o TKI) exhibits diffuse attention and hallucinates pulmonary edema, the complete GDMRG leverages topological priors to focus its attention on the basal and retrocardiac regions. This spatial alignment helps capture concurrent morbidities (e.g., pleural effusion and atelectasis) and suppresses feature hallucinations.}
    \label{fig:case_study}
\end{figure}

\textbf{2) Mechanistic Analysis via Fine-Grained Visual Grounding:} 
To explore the mechanism behind the aforementioned linguistic improvements, we further inspect the fine-grained visual grounding via DGSA heatmaps on a complex patient case presenting with concurrent morbidities (notably \textit{cardiomegaly}, bilateral \textit{pleural effusions}, and basilar \textit{atelectasis}), as depicted in Fig. \ref{fig:case_study}. 

Without explicit topological constraints, the ablated baseline exhibits a more diffuse attention distribution. It struggles to capture the inherent pathological correlations between pleural effusion and atelectasis. This leads to a false positive, generating an anatomically unsupported diagnosis of ``mild-to-moderate pulmonary edema.''

In contrast, modulated by the TKI module, the network incorporates the prior that an enlarged heart frequently co-occurs with basal effusion and volume loss. Guided by this topological prior, the DGSA localizes, recalibrating the visual focus onto the bilateral lung bases and the retrocardiac area. This localized spatial alignment helps suppress feature hallucinations and facilitates the generation of precise clinical terminologies, such as ``\textit{compressive atelectasis}'' and ``\textit{retrocardiac opacification}.'' Overall, this indicates that the TKI module helps bridge macroscopic disease co-occurrences with regional visual grounding, reducing diagnostic omissions and improving vision-language alignment.

\section{Conclusion}
In this paper, we introduced the Graph-Augmented Dual-Stream Medical Report Generation with Topological Internalization (GDMRG) framework to address the challenges of complex disease co-occurrences and long-tailed diagnostic omissions in MRG systems. By introducing the Topological Knowledge Internalization (TKI) module, we transformed implicit clinical complication priors into explicit parameterized constraints. Coupled with a dual-stream classification architecture and a Truncated Asymmetric Loss (T-ASL) optimization strategy, our framework extracts these topological priors as discrete diagnostic prompts. These prompts subsequently guide the Diagnosis-Guided Spatial Attention (DGSA) module, enhancing visual grounding and reducing semantic inconsistencies. Experimental results demonstrate the effectiveness of our proposed paradigm. GDMRG demonstrates improved Clinical Efficacy (CE) on the MIMIC-CXR dataset, particularly in reducing false negatives for sparse and concurrent lesions. While we observe a marginal trade-off in certain N-gram-based linguistic metrics, we consider this a clinically acceptable compromise—prioritizing diagnostic faithfulness over strict syntactic alignment with high-frequency normal templates. Furthermore, the structural constraints provided by the explicit topological graph contribute to the model's zero-shot generalization capabilities, as evidenced by its performance on the out-of-domain IU X-Ray dataset. In conclusion, by establishing a parameterized, interpretable closed loop from macroscopic disease topology to microscopic text decoding, this work presents a clinically relevant framework that can inform the development of future medical AI assistants.

\section*{Supplementary Material}
Sample-level clinical efficacy results, class-specific ROC analysis, implementation notes for the DSE module, further ablation studies (on autoregressive prompt supervision and topological sparsity threshold), and an extended multi-sample qualitative analysis are provided in the appended supplementary material.


\clearpage 

\twocolumn[
  \begin{center}
    \Large \textbf{Supplementary Material for ``Graph-Augmented Topological Internalization with Dual-Stream Classifiers for Medical Report Generation''}
    \vspace{0.8cm}
  \end{center}
]

\setcounter{section}{0}
\renewcommand{\thesection}{S-\Roman{section}}
\setcounter{table}{0}
\renewcommand{\thetable}{S\arabic{table}}
\setcounter{figure}{0}
\renewcommand{\thefigure}{S\arabic{figure}}
\setcounter{equation}{0}
\renewcommand{\theequation}{S\arabic{equation}}

\section{Supplementary Results on Sample-level Clinical Efficacy}
We report the ablation results of GDMRG using the sample-level macro-averaging protocol for clinical efficacy (CE) metrics. As shown in Table \ref{tab:ablation_components_sample}, integrating the OTS, TKI, and DGSA modules consistently improves diagnostic performance. Although the absolute values under this sample-level protocol are naturally lower than their micro-averaged counterparts, the relative performance gains across modules remain robust, validating the proposed architecture under different evaluation metrics.

\begin{table}[htbp]
\centering
\caption{Supplementary ablation study of different network components evaluated under the sample-level CE protocol. \textbf{OTS}: Optimal Threshold Selection; \textbf{DSE}: Diagnostic Semantic Enhancement; \textbf{TKI}: Topological Knowledge Internalization; \textbf{DGSA}: Diagnosis-Guided Spatial Attention.}
\label{tab:ablation_components_sample}
\resizebox{\linewidth}{!}{
\begin{tabular}{lcccc}
\toprule
\textbf{Models} & Precision & Recall & F1 & BLEU-1 \\
\midrule
BASE & 0.500 & 0.424 & 0.433 & 0.388 \\
BASE + OTS & 0.453 & 0.527 & 0.459 & 0.389 \\
BASE + DSE + DGSA & 0.504 & 0.427 & 0.434 & 0.383 \\
BASE + OTS + DSE + DGSA & 0.453 & 0.536 & 0.462 & 0.404 \\
BASE + DSE + TKI & 0.509 & 0.447 & 0.448 & 0.396 \\
BASE + OTS + DSE + TKI & 0.460 & \textbf{0.549} & 0.472 & 0.399 \\
BASE + DSE + TKI + DGSA & \textbf{0.501} & 0.420 & 0.430 & 0.383 \\
BASE + OTS + DSE + TKI + DGSA & 0.473 & 0.546 & \textbf{0.477} & \textbf{0.406} \\
\bottomrule
\end{tabular}%
}
\end{table}

\section{Detailed Receiver Operating Characteristic (ROC) Analysis}
We present the class-specific Receiver Operating Characteristic (ROC) curves and corresponding Area Under the Curve (AUC) values for the 14 core chest abnormalities. 

\begin{figure}[htbp]
    \centering
    \includegraphics[width=\linewidth]{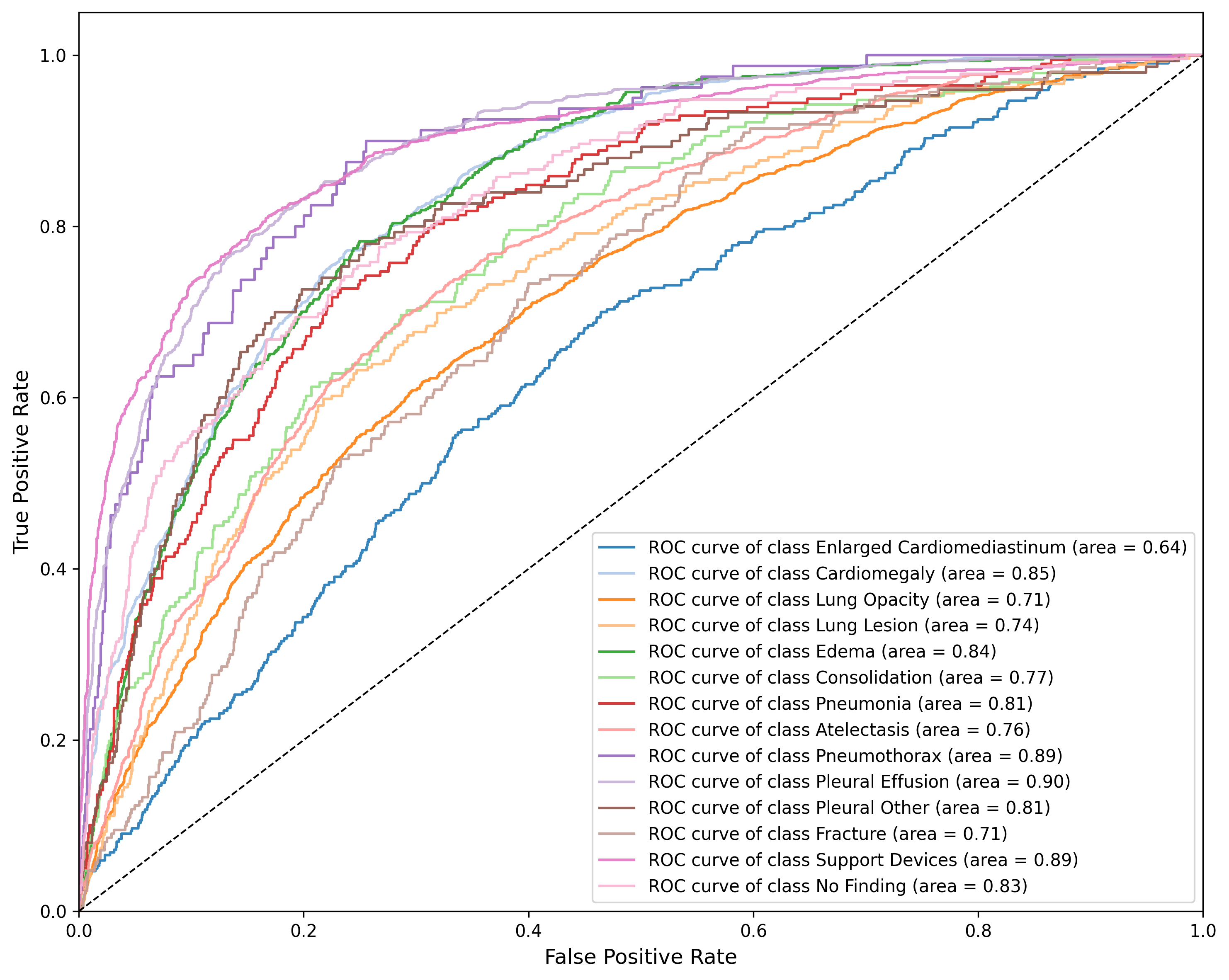} 
    \caption{Class-specific Receiver Operating Characteristic (ROC) curves and Area Under the Curve (AUC) scores for the 14 core diagnostic categories evaluated on the MIMIC-CXR test set.}
    \label{fig:roc_curve}
\end{figure}

As depicted in Fig. \ref{fig:roc_curve}, the classifier demonstrates strong diagnostic performance across most pathological categories. Notably, the model achieves high AUC scores in identifying conditions with strong topological co-occurrences and distinct visual features, such as \textit{Pleural Effusion} (0.90) and \textit{Pneumothorax} (0.89). Furthermore, the \textit{No Finding} category yields an AUC of 0.83, demonstrating effective differentiation of healthy physiological states. 

Conversely, performance on morphologically subtle categories, such as \textit{Enlarged Cardiomediastinum} (0.64), remains constrained. This aligns with clinical consensus, where lesion presentation is highly sensitive to patient positioning and projection angles. Overall, this ROC analysis supports the utility of the dual-stream classification system as an effective diagnostic front-end.

\section{Notes on DSE Module Implementation and Mathematical Equivalence}
Regarding the codebase implementation, the Diagnostic Semantic Enhancement (DSE) module utilizes generalized architectural components (e.g., \textit{EfficientMultiHeadAttention} and \textit{LocalityFeedForward} blocks). 

Within the current GDMRG pipeline, the input to the DSE module is the globally pooled visual representation ($\mathbf{v}_{global} \in \mathbb{R}^{B \times 2048}$). Since the spatial dimension is aggregated ($H=W=1$), the multi-head self-attention mechanism mathematically reduces to a Multi-Layer Perceptron (MLP) mapping, and the localized convolutional operations function as point-wise ($1 \times 1$) linear projections.

Thus, the effective operation corresponds to the theoretical description in Section III-B: a linear feature projection followed by a Squeeze-and-Excitation (SE) channel attention mechanism. We retained this generalized implementation for two primary reasons: 

\begin{itemize}
    \item \textbf{Checkpoint Compatibility:} To ensure strict parametric alignment with pre-trained weights, enabling researchers to evaluate provided checkpoints without architectural mismatch.
    \item \textbf{Architectural Extensibility:} To maintain a forward-compatible interface for future iterations, where unpooled spatial patches could be directly routed for fine-grained multi-instance learning.
\end{itemize}

\section{Ablation on Autoregressive Prompt Supervision}
In conventional prompt-driven generation pipelines, discrete diagnostic prefixes are typically treated as conditional inputs. Consequently, during the training phase, these prompt tokens are often masked within the cross-entropy loss calculation (i.e., the loss is computed solely over the generated report). In our GDMRG framework, we investigated whether requiring the language decoder to autoregressively reconstruct these prompt tokens could yield additional representational benefits.

We conducted an ablation study comparing the standard \textbf{Prompt Masking (PM)} strategy against our adopted \textbf{Full Sequence Supervision} approach. In the latter, the Negative Log-Likelihood loss ($\mathcal{L}_{NLL}$) is calculated across the entire sequence, including the prefix prompt tokens. The quantitative results are detailed in Table \ref{tab:ablation_prompt_mask}.

As shown, removing the prompt mask and applying full sequence supervision yields a modest improvement across both Clinical Efficacy (F1: 0.556 $\rightarrow$ 0.561) and Natural Language Generation metrics (BLEU-1: 0.400 $\rightarrow$ 0.406). We posit that requiring the text decoder to explicitly predict the diagnostic concepts acts as an auxiliary semantic regularization. This mechanism encourages the earlier layers of the decoder to achieve a tighter alignment with the fused multimodal context ($\mathbf{V}_{fused}$) before autoregressively decoding the clinical narrative. Ultimately, this sequential reinforcement contributes to both the diagnostic accuracy and the textual coherence of the generated reports.

\begin{table}[htbp] 
\centering
\caption{Ablation study on the training supervision strategy for diagnostic prompts. \textbf{Prompt Masking (PM)} excludes the prefix tokens from the loss calculation. The default GDMRG utilizes \textbf{Full Sequence Supervision}, applying the autoregressive loss to both the prompt and the generated report.}
\label{tab:ablation_prompt_mask}
\resizebox{\linewidth}{!}{
\begin{tabular}{lcccc}
\toprule
\textbf{Supervision Strategy} & Precision & Recall & F1 & BLEU-1 \\
\midrule
GDMRG w/ Prompt Masking (PM) & 0.510 & \textbf{0.612} & 0.556 & 0.400 \\
GDMRG (Full Sequence Supervision) & \textbf{0.519} & 0.610 & \textbf{0.561} & \textbf{0.406} \\
\bottomrule
\end{tabular}%
}
\end{table}

\begin{figure*}[htbp] 
    \centering
    \includegraphics[width=\textwidth, keepaspectratio]{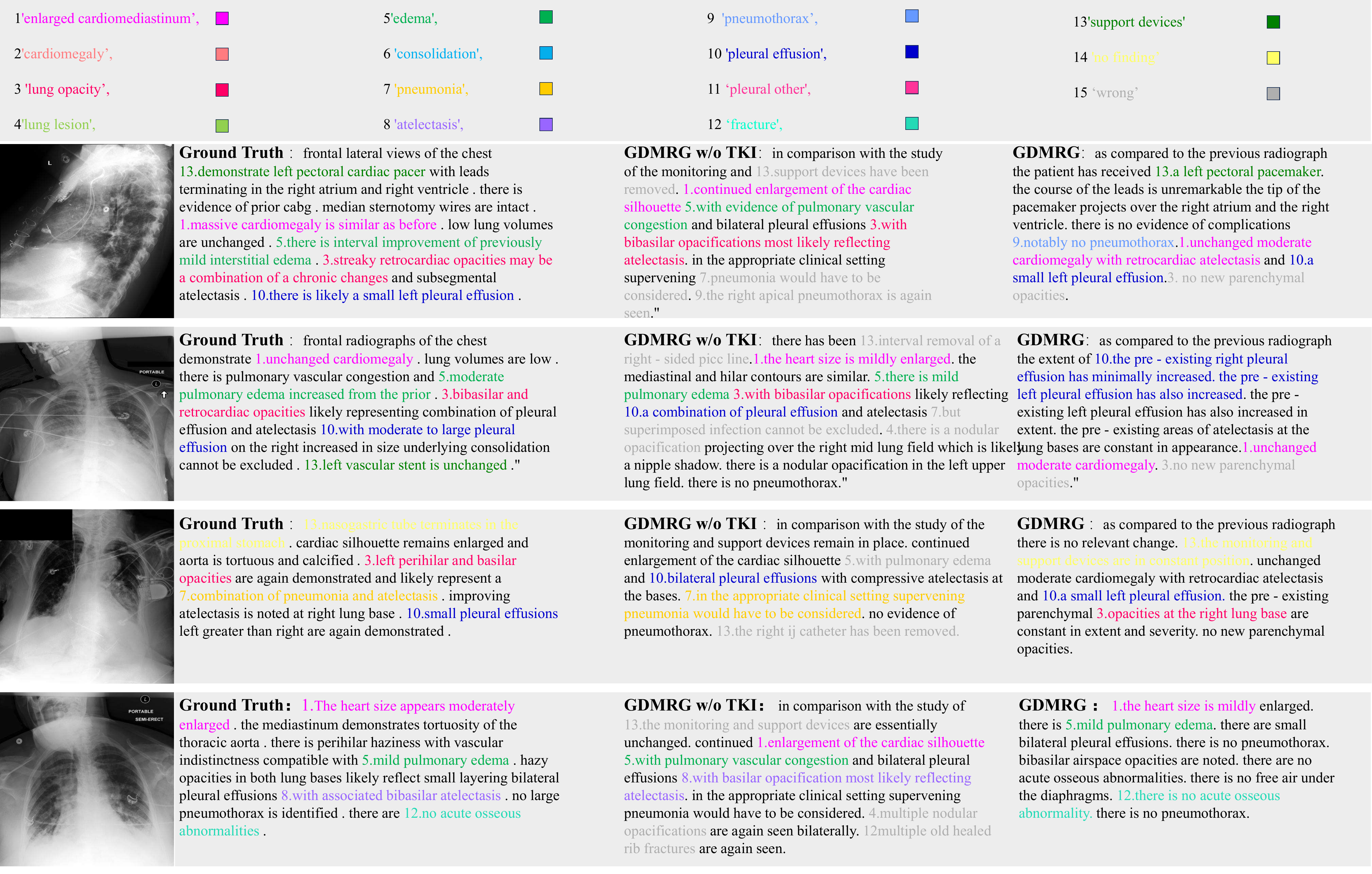} 
    \caption{Multi-sample qualitative comparison between the ablated baseline (w/o TKI) and the complete GDMRG. The complete GDMRG effectively suppresses clinical hallucinations (e.g., false positive fractures) and disentangles complex disease co-occurrences. The explicit topological priors ensure both semantic coherence and logical consistency in auxiliary attributes.}
    \label{fig:multi_case_study_supp}
\end{figure*}

\section{Sensitivity Analysis of the Topological Sparsity Threshold}
To determine the optimal graph structure for the Topological Knowledge Internalization (TKI) module, we conducted a hyperparameter sensitivity analysis on the topological sparsity threshold, denoted as $\phi$. This threshold dictates the percentile of edges retained in the label co-occurrence adjacency matrix, effectively filtering out low-frequency, potentially spurious disease correlations. The quantitative results are presented in Table \ref{tab:sensitivity_sparsity}.

As shown in the table, setting a lower threshold (e.g., $\phi=70$ or $80$, retaining the top 30\% or 20\% of edges respectively) results in an overly dense graph. While this preserves a relatively high Recall, it introduces substantial topological noise by linking weakly correlated or physically disjoint conditions. This over-smoothing effect misguides the classifier, leading to a noticeable degradation in Precision (0.449 for $\phi=70$) and the overall F1-score.

Conversely, applying an overly strict threshold (e.g., $\phi=95$, retaining only the top 5\% of edges) creates a highly sparse graph. Although this mitigates noise, it aggressively discards critical, albeit less frequent, clinical comorbidities. As a result, the network loses valuable topological guidance for complex concurrent lesions, causing a drop in both Recall and F1-score compared to the optimal setting.

Ultimately, our default configuration of $\phi=90$ (retaining the top 10\% of edges) achieves the optimal balance. It successfully filters out noisy statistical correlations while preserving essential clinical comorbidity hubs, thereby yielding the highest comprehensive performance in both diagnostic accuracy (F1: 0.561) and natural language generation quality (BLEU-1: 0.406).

\begin{table}[htbp] 
\centering
\caption{Sensitivity analysis of the topological sparsity threshold $\phi$ on the MIMIC-CXR dataset. $\phi$ dictates the percentile of edge retention in the label co-occurrence graph. Setting a lower threshold (e.g., $\phi=70$, retaining top 30\% of edges) introduces spurious correlations as topological noise, whereas an overly strict threshold ($\phi=95$, retaining top 5\%) discards critical clinical comorbidities. Our default configuration ($\phi=90$, retaining top 10\%) achieves the optimal balance between topological knowledge injection and noise suppression.}
\label{tab:sensitivity_sparsity}
\resizebox{\linewidth}{!}{
\begin{tabular}{lcccc}
\toprule
\textbf{Model Variant} & Precision & Recall & F1 & BLEU-1 \\
\midrule
GDMRG-Top30\% ($\phi=70$) & 0.449 & 0.606 & 0.515 & 0.384 \\
GDMRG-Top20\% ($\phi=80$) & 0.488 & 0.574 & 0.528 & 0.394 \\
GDMRG-Top5\% ($\phi=95$)  & 0.500 & 0.589 & 0.541 & 0.400 \\
\textbf{GDMRG-Top10\% ($\phi=90$, Ours)} & \textbf{0.519} & \textbf{0.610} & \textbf{0.561} & \textbf{0.406} \\
\bottomrule
\end{tabular}%
}
\end{table}

\section{Extended Multi-Sample Qualitative Analysis}
In addition to the fine-grained visual grounding presented in the main text, we provide an extended multi-sample textual comparison to demonstrate linguistic coherence and hallucination suppression across diverse clinical scenarios (Fig. \ref{fig:multi_case_study_supp}). 

We compare the generated reports of our baseline model (GDMRG w/o TKI) against the complete GDMRG. In the absence of explicit complication priors, the baseline model frequently exhibits limitations when handling long-tail or visually ambiguous entities, leading to clinical hallucinations. For instance, the baseline generates structurally independent abnormalities without sufficient visual evidence, such as "multiple old healed rib fractures," while generating unsupported hedging clauses like "superimposed infection cannot be excluded." Conversely, guided by the learned topological graph, GDMRG mitigates the generation of isolated orthopaedic findings that lack statistical and anatomical support in the given clinical context, thereby yielding a clinically consistent description ("no new parenchymal opacities").

Furthermore, the baseline exhibits difficulty in reflecting complex clinical associations, often misattributing opacities entirely to isolated atelectasis when cardiomegaly and pleural effusions are concurrent. In contrast, guided by the parameterized weights from the TKI module, GDMRG incorporates this clinical triad, generating sequences that describe the concurrent conditions: "unchanged moderate cardiomegaly... the pre-existing left pleural effusion has also increased." Additionally, GDMRG avoids contradictory statements regarding auxiliary attributes (e.g., support devices), generating sequences that better align with the ground-truth reference reports.


\begin{thebibliography}{100}

\bibitem{jing2018automatic} B. Jing, P. Xie, and E. Xing, ``On the automatic generation of medical imaging reports,'' in \textit{Proceedings of ACL}, 2018, pp. 2577--2586.
\bibitem{chen2020generating} Z. Chen, Y. Song, T. H. Chang, and X. Wan, ``Generating radiology reports via memory-driven transformer,'' in \textit{Proceedings of EMNLP}, 2020, pp. 1439--1449.
\bibitem{brady2017error} A. Brady, ``Error and discrepancy in radiology: Inevitable or avoidable?'' \textit{Insights into Imaging}, vol. 8, no. 1, pp. 171--182, 2017.
\bibitem{yu2020c2fnas} Q.~Yu, D.~Yang, H.~Roth, Y.~Bai, Y.~Zhang, A.~L. Yuille, and D.~Xu, ``C2FNAS: Coarse-to-fine neural architecture search for 3D medical image segmentation,'' in \textit{Proceedings of the IEEE/CVF Conference on Computer Vision and Pattern Recognition (CVPR)}, 2020, pp. 4126--4135.
\bibitem{liu2019clinically} G. Liu et al., ``Clinically accurate chest X-ray report generation,'' in \textit{Proceedings of MLHC}, 2019, pp. 249--269.
\bibitem{chexgcn}
H.~Chen, S.~Miao, D.~Xu, G.~D.~Hager, and A.~P.~Harrison, ``Label co-occurrence learning with graph convolutional networks for multi-label chest X-ray image classification,'' \textit{IEEE J. Biomed. Health Inform.}, vol. 24, no. 8, pp. 2292--2302, 2020.
\bibitem{pennington2014glove} J. Pennington, R. Socher, and C. D. Manning, ``GloVe: Global vectors for word representation,'' in \textit{Proceedings of EMNLP}, 2014, pp. 1532--1543.
\bibitem{zhou2022learning} K. Zhou et al., ``Learning to prompt for vision-language models,'' \textit{International Journal of Computer Vision}, vol. 130, no. 9, pp. 2337--2348, 2022.
\bibitem{miura2021improving} Y. Miura, Y. Zhang, E.  Tsai, C. Langlotz, and D. Jurafsky, ``Improving factual completeness and consistency of image-to-text radiology report generation,'' in \textit{Proceedings of NAACL}, 2021, pp. 5288--5304.
\bibitem{cui2019class} Y. Cui, M. Jia, T. Lin, Y. Song, and S. Belongie, ``Class-balanced loss based on effective number of samples,'' in \textit{Proceedings of CVPR}, 2019, pp. 9268--9277.
\bibitem{ridnik2021asymmetric} T. Ridnik, E. Ben-Baruch, N. Zamir, A. Noy, I. Friedman, M. Protter, and L. Zelnik-Manor, ``Asymmetric loss for multi-label classification,'' in \textit{Proceedings of ICCV}, 2021, pp. 82--91.
\bibitem{wang2018tienet} X. Wang, Y. Peng, L. Lu, Z. Lu, M. Bagheri, and R. M. Summers, ``TieNet: Text-image embedding network for common thorax disease classification and reporting in chest X-rays,'' in \textit{Proceedings of CVPR}, 2018, pp. 9049--9058.
\bibitem{xue2018multimodal} Y. Xue et al., ``Multimodal recurrent model with attention for automated radiology report generation,'' in \textit{Proceedings of MICCAI}, 2018, pp. 457--466.
\bibitem{li2018hybrid} Y. Li, X. Liang, Z. Hu, and E. P. Xing, ``Hybrid retrieval-generation reinforced agent for medical image report generation,'' in \textit{Advances in Neural Information Processing Systems}, 2018.
\bibitem{rui2021artificial} E. J. Topol, ``High-performance medicine: The convergence of human and artificial intelligence,'' \textit{Nature Medicine}, vol. 25, no. 1, pp. 44--56, 2019.
\bibitem{promptmrg} H. Jin, H. Che, Y. Lin, H. Chen et al., ``PromptMRG: Diagnosis-driven prompts for medical report generation,'' in \textit{Proceedings of AAAI}, 2024, pp. 15432--15440.
\bibitem{jia2021visual} C. Jia et al., ``Scaling up visual and vision-language representation learning with noisy text supervision,'' in \textit{Proceedings of ICML}, 2021, pp. 4904--4916.
\bibitem{liu2023pre} P. Liu et al., ``Pre-train, prompt, and predict: A systematic survey of prompting methods in natural language processing,'' \textit{ACM Computing Surveys}, vol. 55, no. 9, pp. 1--35, 2023.
\bibitem{chen2021cross} Z. Chen, Y. Shen, Y. Song, and X. Wan, ``Cross-modal memory networks for radiology report generation,'' in \textit{Proceedings of ACL}, 2021, pp. 5904--5914.
\bibitem{guan2021radgraph} S. Jain et al., ``RadGraph: Extracting clinical entities and relations from radiology reports,'' in \textit{Advances in Neural Information Processing Systems}, 2021.
\bibitem{kipf2016semi} T. N. Kipf and M. Welling, ``Semi-supervised classification with graph convolutional networks,'' in \textit{Proceedings of ICLR}, 2017.
\bibitem{velickovic2017graph} P. Veli\v{c}kovi\'{c} et al., ``Graph attention networks,'' in \textit{Proceedings of ICLR}, 2018.
\bibitem{hou2021ratchet} B. Hou et al., ``Ratchet: Medical transformer for chest X-ray diagnosis and reporting,'' in \textit{Proceedings of MICCAI}, 2021, pp. 293--303.
\bibitem{zhang2020when} Y. Zhang et al., ``When radiology report generation meets knowledge graph,'' in \textit{Proceedings of AAAI}, 2020, pp. 12910--12917.
\bibitem{endo2021retrieval} M. Endo et al., ``Retrieval-based chest X-ray report generation using a pre-trained contrastive language-image model,'' in \textit{Proceedings of MLHC}, 2021, pp. 209--230.
\bibitem{zhang2023knowledge}
X.~Zhang, C.~Wu, Z.~Zhao, W.~Lin, Y.~Zhang, Y.~Wang, and W.~Xie, ``Knowledge-enhanced visual-language pre-training on chest radiology images,'' \textit{Nature Communications}, vol. 14, no. 1, p. 4542, 2023.
\bibitem{cao2019learning} K. Cao, C. Wei, A. Gaidon, N. Arechiga, and T. Ma, ``Learning imbalanced datasets with label-distribution-aware margin loss,'' in \textit{Advances in Neural Information Processing Systems}, 2019.
\bibitem{chawla2002smote} N. V. Chawla, K. W. Bowyer, L. O. Hall, and W. P. Kegelmeyer, ``SMOTE: Synthetic minority over-sampling technique,'' \textit{Journal of Artificial Intelligence Research}, vol. 16, pp. 321--357, 2002.
\bibitem{lin2017focal} T.-Y. Lin, P. Goyal, R. Girshick, K. He, and P. Doll\'{a}r, ``Focal loss for dense object detection,'' in \textit{Proceedings of ICCV}, 2017, pp. 2980--2988.
\bibitem{kang2019decoupling} B. Kang, S. Xie, M. Rohrbach, Z. Yan, A. Gordo, J. Feng, and Y. Kalantidis, ``Decoupling representation and classifier for long-tailed recognition,'' in \textit{Proceedings of ICLR}, 2020.
\bibitem{menon2020long} A. K. Menon, S. Jayasumana, A. S. Rawat, H. Jain, A. Veit, and S. Kumar, ``Long-tail learning via logit adjustment,'' in \textit{Proceedings of ICLR}, 2021.
\bibitem{guo2017calibration} C. Guo, G. Pleiss, Y. Sun, and K. Q. Weinberger, ``On calibration of modern neural networks,'' in \textit{Proceedings of ICML}, 2017, pp. 1321--1330.
\bibitem{he2016deep} K. He, X. Zhang, S. Ren, and J. Sun, ``Deep residual learning for image recognition,'' in \textit{Proceedings of CVPR}, 2016, pp. 770--778.
\bibitem{hu2018squeeze} J. Hu, L. Shen, and G. Sun, ``Squeeze-and-excitation networks,'' in \textit{Proceedings of CVPR}, 2018, pp. 7132--7141.
\bibitem{vaswani2017attention} A. Vaswani et al., ``Attention is all you need,'' in \textit{Advances in Neural Information Processing Systems}, 2017.
\bibitem{devlin2018bert} J. Devlin, M. Chang, K. Lee, and K. Toutanova, ``BERT: Pre-training of deep bidirectional transformers for language understanding,'' in \textit{Proceedings of NAACL}, 2019, pp. 4171--4186.
\bibitem{loshchilov2017decoupled} I. Loshchilov and F. Hutter, ``Decoupled weight decay regularization,'' in \textit{Proceedings of ICLR}, 2019.
\bibitem{johnson2019mimic} A. E. W. Johnson et al., ``MIMIC-CXR, a de-identified publicly available database of chest radiographs with free-text reports,'' \textit{Scientific Data}, vol. 6, no. 1, p. 317, 2019.
\bibitem{demner2016preparing} D. Demner-Fushman et al., ``Preparing a collection of radiology examinations for distribution and retrieval,'' \textit{J. Am. Med. Inform. Assoc.}, vol. 23, no. 2, pp. 304--310, 2016.
\bibitem{irvin2019chexpert} J. Irvin et al., ``CheXpert: A large chest radiograph dataset with uncertainty labels and expert comparison,'' in \textit{Proceedings of AAAI}, 2019, pp. 590--597.
\bibitem{papineni2002bleu} K. Papineni, S. Roukos, T. Ward, and W. J. Zhu, ``BLEU: A method for automatic evaluation of machine translation,'' in \textit{Proceedings of ACL}, 2002, pp. 311--318.
\bibitem{banerjee2005meteor} S. Banerjee and A. Lavie, ``METEOR: An automatic metric for MT evaluation with improved correlation with human judgments,'' in \textit{Proceedings of ACL}, 2005, pp. 65--72.
\bibitem{lin2004rouge} C.-Y. Lin, ``ROUGE: A package for automatic evaluation of summaries,'' in \textit{Proceedings of ACL}, 2004.
\bibitem{m2tr} F. Nooralahzadeh, N. P. Gonzalez, T. Frauenfelder, K. Fujimoto, and M. Krauthammer, ``Progressive transformer-based generation of radiology reports,'' in \textit{Findings of the Association for Computational Linguistics: EMNLP}, 2021, pp. 2824--2832.
\bibitem{liu2021contrastive} F.~Liu, C.~Yin, X.~Wu, S.~Ge, P.~Zhang, and X.~Sun, ``Contrastive attention for automatic chest X-ray report generation,'' in \textit{Findings of the Association for Computational Linguistics: ACL-IJCNLP 2021}, 2021, pp. 269--280.
\bibitem{cvt2dis} A. Nicolson, J. Dowling, and B. Koopman, ``Improving chest X-ray report generation by leveraging warm-starting,'' \textit{Artif. Intell. Med.}, vol. 144, p. 102633, 2023.
\bibitem{m2kt} S. Yang, X. Wu, S. Ge, Z. Zheng, S. K. Zhou, and L. Xiao, ``Radiology report generation with a learned knowledge base and multi-modal alignment,'' \textit{Medical Image Anal.}, vol. 86, p. 102798, 2023.
\bibitem{metrans} Z. Wang, L. Liu, L. Wang, and L. Zhou, ``ME Transformer: Radiology report generation by transformer with multiple learnable expert tokens,'' in \textit{Proceedings of CVPR}, 2023, pp. 11558--11567.
\bibitem{dcl} M. Li, B. Lin, Z. Chen, H. Lin, X. Liang, and X. Chang, ``Dynamic graph enhanced contrastive learning for chest X-ray report generation,'' in \textit{Proceedings of CVPR}, 2023, pp. 3334--3343.
\bibitem{rgrg} T.~Tanida, P.~M\"{u}ller, G.~Kaissis, and D.~Rueckert, ``Interactive and explainable region-guided radiology report generation,'' in \textit{Proceedings of the IEEE/CVF Conference on Computer Vision and Pattern Recognition (CVPR)}, 2023, pp. 7433--7442.
\bibitem{camanet} J. Wang, A. Bhalerao, T. Yin, S. See, and Y. He, ``CAMANet: Class activation map guided attention network for radiology report generation,'' \textit{IEEE J. Biomed. Health Inform.}, vol. 28, no. 4, pp. 2199--2210, 2024.
\bibitem{teaser} J. Zhao, Y. Zhou, Z. Chen, H. Fu, and L. Wan, ``Topicwise separable sentence retrieval for medical report generation,'' \textit{IEEE Trans. Med. Imaging}, vol. 44, no. 3, pp. 1505--1514, 2025.
\bibitem{cmcrl} W. Chen et al., ``Cross-modal causal representation learning for radiology report generation,'' \textit{IEEE Trans. Image Process.}, vol. 34, pp. 2970--2985, 2025.

\end{thebibliography}
\end{document}